\documentclass[runningheads]{llncs}
\usepackage{graphicx}
\usepackage{amsmath,amssymb} 
\usepackage{color}

\begin{document}

\title{3D Pick \& Mix: Object Part Blending in Joint Shape and Image Manifolds}
\titlerunning{3D Pick \& Mix: Object Part Blending in Joint Shape and Image Manifolds} 


\author{Adrian Penate-Sanchez\inst{1,2}\orcidID{0000-0003-2876-3301} \\
Lourdes Agapito\inst{1}\orcidID{0000-0002-6947-1092} }
%

\authorrunning{A. Penate-Sanchez, L. Agapito} 


\institute{University College London \\
Department of Computer Science, United Kingdom \and
University of Oxford \\
Oxford Robotics Institute, United Kingdom \\
\email{adrian@robots.ox.ac.uk\thanks{This work was supported by the SecondHands project, funded from the EU Horizon 2020 Research and Innovation programme under grant agreement 643950 and by the EPSRC grants RAIN and ORCA (EP/R026084/1, EP/R026173/1).}}}

\maketitle

\begin{abstract}
We present {\bf 3D Pick \& Mix,} a new 3D shape retrieval system that
provides users with a new level of freedom to explore 3D shape and
Internet image  collections by introducing the ability to reason about
objects at the level of their constituent parts. While classic
retrieval systems can only formulate simple searches such as
\emph{``find the 3D model that is most similar to the input image''}
our new approach can formulate advanced and semantically meaningful
search queries such as: \emph{``find me the 3D model that best 
combines the design of the legs of the chair in image 1 but with 
no armrests, like the chair in image 2''}. Many applications could benefit 
from such rich queries, users could browse through
catalogues of furniture and \textbf{pick} and \textbf{mix} parts,
combining for example the legs of a chair from one shop
and the armrests from another shop.

\end{abstract}


\section{Introduction}

\begin{figure*}[t!]
  \centering
  \includegraphics[width=\textwidth, trim=0mm 0mm 0mm 0mm]{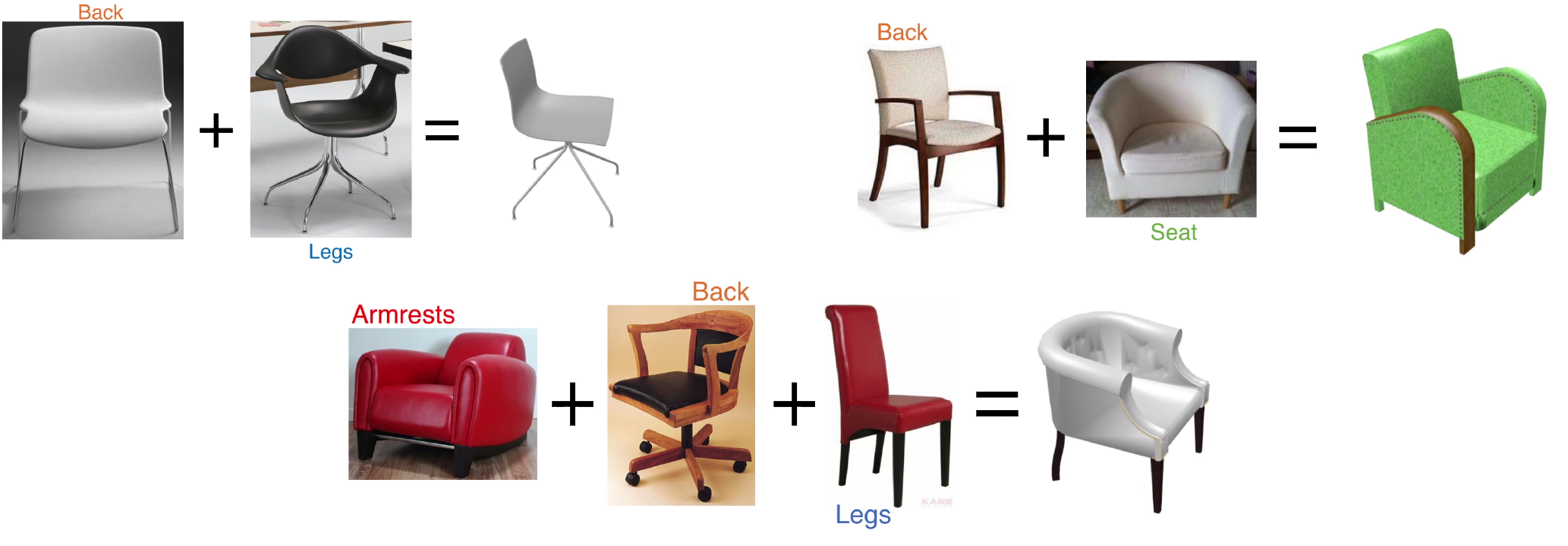}
    \caption{ Our approach takes two (or more) inputs, either
      RGB images or 3D models, and part labels such as
      \emph{"legs"} or \emph{"backrest"} and retrieves a shape that
      combines object parts or properties from all inputs, via a
      cross-manifold optimization technique.}
    \label{fig:teaser}
\end{figure*}

As databases of images and 3D shapes keep growing in size and number,
organizing and exploring them has become increasingly complex. While
most tools so far have dealt with shape and appearance modalities
separately, some recent
methods~\cite{li2015jointembedding,hueting2015} have begun to exploit
the complementary nature of these two sources of information and to
reap the benefits of creating a common representation for images and
3D models. Once images and 3D shapes are linked together, many
possibilities open up to transfer what is learnt from one modality to
another. Creating a joint embedding allows to retrieve 3D models based
on image queries (or vice-versa) or to align images of similar 3D
shapes. However, recent retrieval methods still fall short of being
flexible enough to allow advanced queries. Crucially, they are limited
to reasoning about objects as a whole -- taking a single query image
(or shape) as input at test time prevents them from combining object
properties from different inputs. {\bf 3D Pick \& Mix} overcomes this
limitation by reasoning about objects at the level of parts. It can
formulate advanced queries such as: \emph{``find me the 3D model that
best combines the design of the backrest of the chair in image $1$
with the shape of the legs of the chair in image $2$''} (see Fig.~\ref{fig:teaser}) or \emph{``retrieve chairs with wheels''}.
The ability to reason at the level of parts provides users with a new
level of freedom to explore 3D shape and image datasets. Users could
browse through catalogues of furniture and \textbf{pick} and
\textbf{mix} parts, combining for example the legs of a favourite chair
from one catalogue and the armrests from another (see Fig.~\ref{fig:teaser}).

\begin{figure*}[t!]
    \centering
    \includegraphics[width=\textwidth]{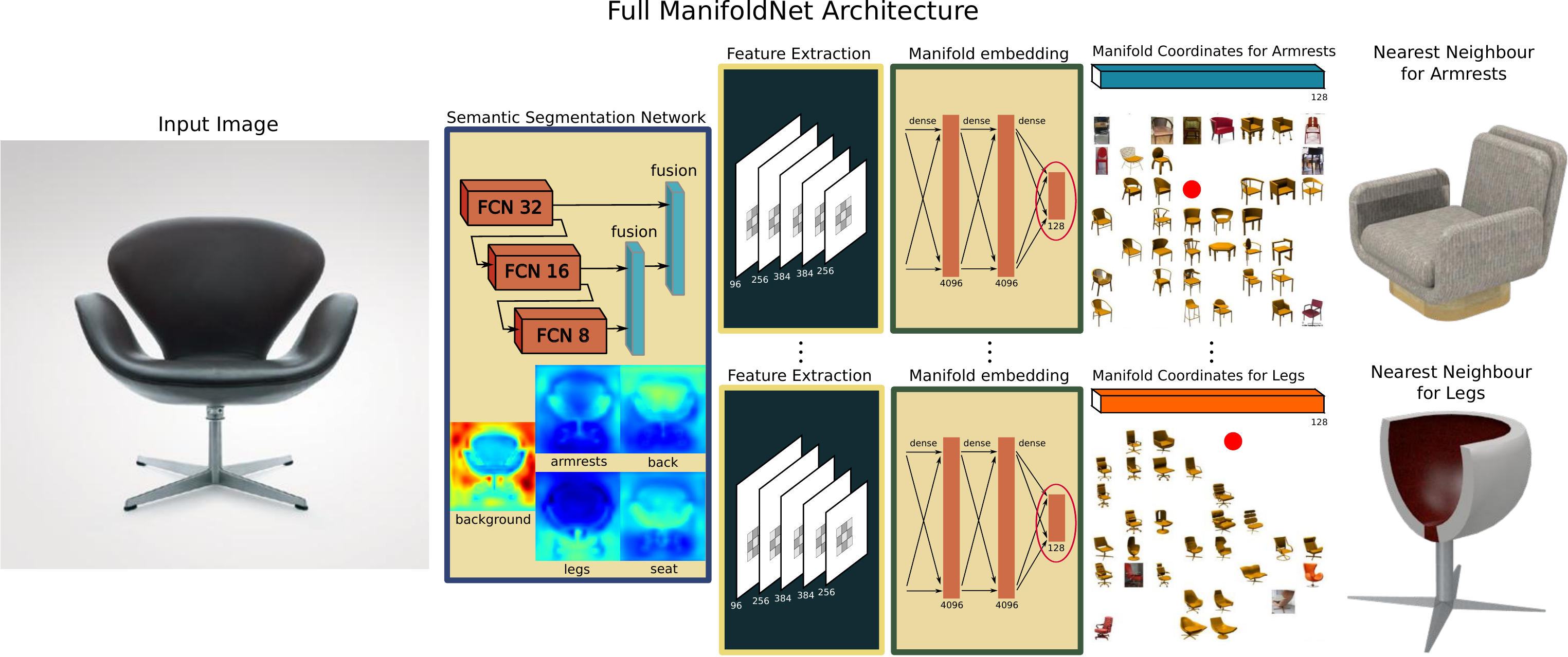}
    \caption{Summary of the architecture of {\bf ManifoldNet}, our new deep network that takes an image as input and learns to regress the coordinates of each object part in the different part manifolds. The architecture has 3 sections: the first set of layers performs semantic segmentation of the image pixels into different semantic parts (such as \emph{''backrest'', ``seat'', ``armrests'' or ``legs''} in the case of chairs). The second section learns an intermediate feature representation for manifold coordinate regression. The final section learns to regress the shape coordinates in each of the part manifolds. We show the nearest neighbour shapes found on the \emph{``armrests''} and \emph{``legs''} manifolds for the depicted input image.}
    \label{fig:manifold_network}
\end{figure*}

Our system first builds independent manifold shape spaces for each
object part (see Fig.~\ref{fig:part_manifold}). A CNN-based manifold
coordinate regressor is trained to map real images of an object to the
part manifolds. Our novel deep architecture jointly performs semantic
segmentation of object parts and coordinate regression for each part
on the corresponding shape manifold. This network is trained using
only synthetic data, Fig.~\ref{fig:manifold_network} illustrates the
architecture. At test time the user provides two (or more)
images (or 3D models) as input and determines which parts they would
like to pick from each (note that this only requires a label name such
as \emph{'legs'}). Our system retrieves the model that best fits the
arrangement of parts by performing a cross-manifold optimization (see
Fig.~\ref{fig:teaser}). The main {\bf contributions} of our 3D Pick \& Mix system are:

\begin{itemize}
\item We learn embeddings (manifolds) for object parts (for instance
  the legs or the armrests of chairs) which allow us to retrieve
  images or 3D models of objects with similarly shaped parts.
\item We propose a new deep architecture that can map RGB images onto
  these manifolds of parts by regressing their coordinates. Crucially,
  the input to the network is simply an RGB image and the name (label)
  of the object part. The CNN learns to: \emph{(i)} segment the pixels that
  correspond to the chosen  part, and \emph{(ii)} regress its coordinates on
  the shape manifold.
\item At query time our retrieval system can combine object parts from
  multiple input images, enabled by a cross-manifold optimization
  technique.
\end{itemize}

\section{Related Work}



\noindent{\bf Joint 3D model/image embeddings:} While most shape retrieval
methods had traditionally dealt with shape and appearance modalities
separately, a recent trend has emerged that exploits the complementary
nature of appearance and shape by creating a common representation for
images and 3D models. ~\cite{hueting2015} exploits the different
advantages of shape and images by using the robustness of 3D models
for alignment and pose estimation and the reliability of image labels
to identify the objects. While they do not explicitly create a joint
manifold based on shape similarity they do rely on image
representations for both modalities. Another example of 3D model/image
embedding is~\cite{li2015jointembedding} who first builds a manifold
representation of 3D shapes and then trains a CNN to recognize the
shape of the object in an image. Unlike our approach,
both~\cite{hueting2015,li2015jointembedding}, limit their
representations to objects as a whole preventing the combination of
properties taken from different inputs.~\cite{Tasse_ToG16} perform
shape retrieval from sketches, words, depth maps and real images by
creating a manifold space that combines the different inputs. Since
intra-class similarity is not the main focus, most instances of the
same class tend to appear clustered.  ~\cite{Lim:2016:StyleLearning}
learn a manifold-space metric by using triplets of shapes where the
first is similar to the third but dissimilar to the second. Similarly to our approach, the metric
space is defined based on shape and not image similarity.
~\cite{Girdhar16b} first generates voxel representations of the
objects present in the RGB image inputs. A shared latent shape
representation is then learnt for both images and the voxelized
data. At test time RGB convolutions and volume generation
deconvolution layers are used to produce the 3D shape. 
%


\noindent{\bf 3D shape blending/mixing:} Much in the line of the work presented
in this paper, there has been fruitful research in shape blending in
recent years.
The \emph{``3D model evolution``} approach
of~\cite{Xu_genetic_ToG2012} takes a small set of 3D models as input
to generate many. Parts from two models cross-over to form a new 3D
model, continuing to merge original models with new ones to generate a
large number of 3D models.
In ~\cite{Alhashim_ToG14} new shapes are generated by interpolating and
varying the topology between two 3D models.
The photo-inspired 3D modeling method of~\cite{Xu_ToG_2011} takes a
single image as input, segments it into parts using an interactive
model-driven approach, then retrieves a 3D model candidate that is
finally deformed to match the silhouette of the input photo.
The probabilistic approach of~\cite{Kalogerakis_ToG_2012} learns a model
that describes the relationship between the parts of 3D shapes which
then allows to create an immense variety of new blended shapes by
mixing attributes from different models.
The sketch driven method of~\cite{Sketch2Design} edits a pre-
segmented 3D shape using user-drawn sketches of new parts. The sketch
is used to retrieve a matching 3D part from a catalogue of 3D shapes
which is then snapped onto the original 3D shape to create a new
blended 3D shape.
Note that the above approaches use only 3D shapes as input for shape
blending, with the exception of~\cite{Xu_ToG_2011} who use a single
photograph and~\cite{Sketch2Design} who use sketches. However, unlike
ours, neither of these approaches can combine different input images
to retrieve a shape that blends parts from each input.


\noindent{\bf Modeling of 3D object parts:} We will differentiate between 3D segmentation approaches that seek to
ensure consistency in the resulting segmentation across different
examples of the same object class (co-segmentation) and those that
seek a semantically meaningful segmentation (semantic segmentation).
Some recent examples of approaches that perform co-segmentation can 
be found in ~\cite{abstractionTulsiani17,Fish_ToG_2016}, but as we seek
to describe parts that have meaning to humans we will focus on the later.
We can find examples of semantic 3D parts in approaches like~\cite{Yi16}. 
~\cite{Yi16} provides accurate semantic region annotations for large geometric datasets with a
fraction of the effort by alternating between using few manual annotations from an expert 
and a system that propagates labels to new models. We exploit the ShapeNet annotations provided by~\cite{Yi16} 
as the ground truth part shape when constructing our joint manifold.


\noindent{\bf Recognition of 3D structure from images:} 
The exemplar-based approach of~\cite{Aubry14} performs joint object
category detection viewpoint estimation, exploiting 3D model datasets
to render instances from different viewpoints and then learn the
combination of viewpoint-instance using exemplar SVMs.
~\cite{choy20163d} uses 3D Convolutional LSTMs to extract the 3D shape
of an object from one or more viewpoints. By using LSTM blocks that
contain memory, they progressively refine the shape of the
object. 
~\cite{Haoqiang_CVPR17} learn to generate a 3D point cloud from a
single RGB image, it learns purely from synthetic data. By using a
point cloud instead of a volumetric representation better definition
of the details of the shape are obtained. Their novel approach learns
how to generate several plausible 3D reconstructions from a single RGB
image at test time if the partial observation of the image is
ambiguous.
~\cite{Su_2015_ICCV} learn to recognize the object category and the
camera viewpoint for an image using synthetically generated images for
training. This work showed that datasets of real images annotated with
3D information were not required to learn  shape properties from 
images as this could be learnt from synthetically generated
renderings.
~\cite{shmlg_imageDepth_sig14} obtain good depth estimates for an
image given a set of 3D models of the same class.  
%

\section{Overview}
\label{sec:outline}

\begin{figure}[t]
    \centering
    \includegraphics[width=\textwidth]{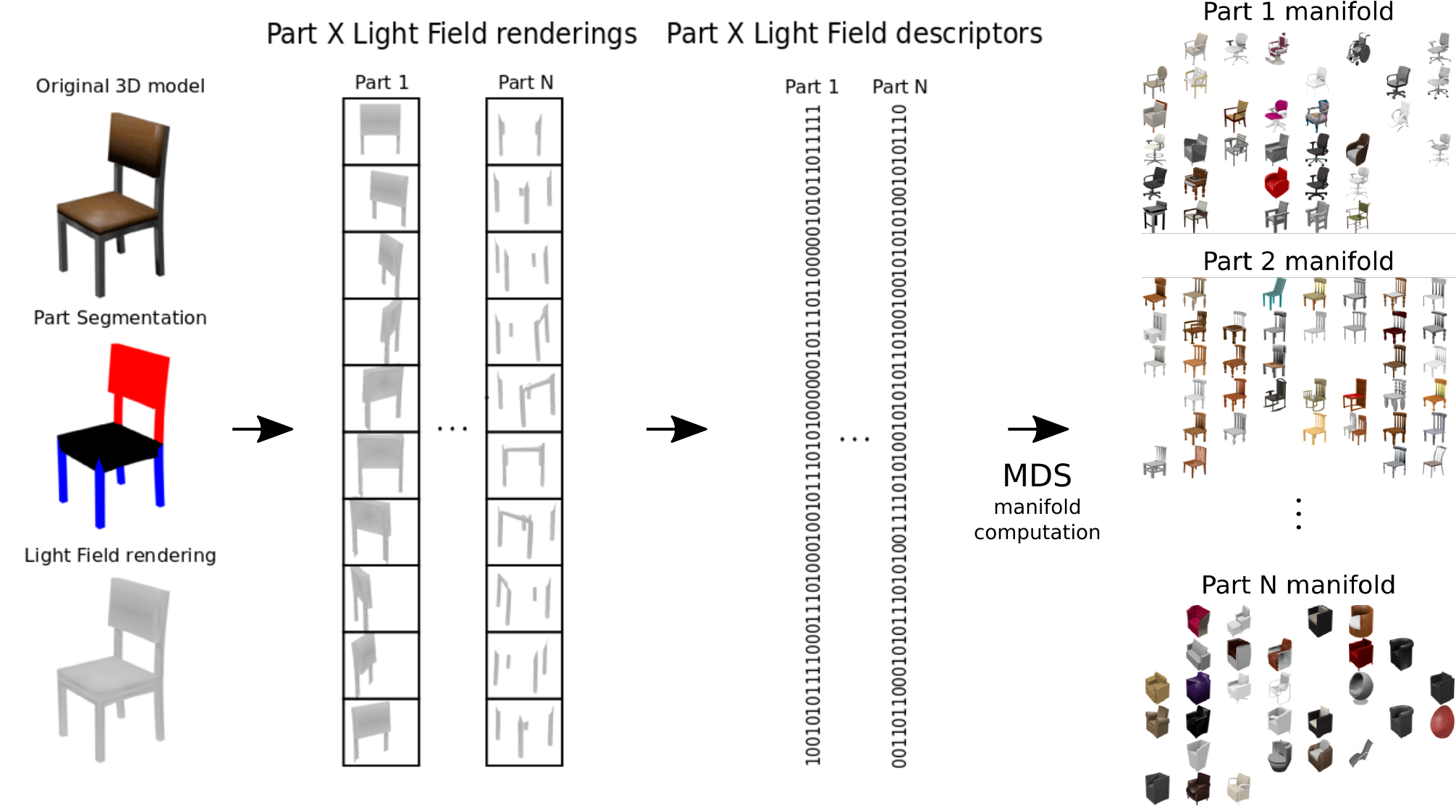}
    \caption{ Shape manifold construction. The shape
      of each part for each 3D model is rendered from different
      viewpoints and represented with a Light Field
      descriptor~\cite{LfDescriptor}. The manifolds are then built using
      non-linear multi-dimensional-scaling (MDS) and the $L_2$ norm
      between feature vectors as the distance metric. In each
      low-dimensional manifold, objects that have similarly
      shaped parts appear close to each other.}
    \label{fig:part_manifold}
\end{figure}

In this section we provide a high level overview of our {\bf 3D Pick \& Mix}
retrieval system.  Our system requires a training stage in which:
\emph{(i)} manifolds of 3D shapes of object parts are built (see
Fig.~\ref{fig:part_manifold}) and \emph{(ii)} a CNN is trained to take
as input an image and regress the coordinates of each of its
constituent parts on the shape manifolds (illustrated in
Fig.~\ref{fig:manifold_network}). At query time the system receives an
image or set of images as input and obtains the corresponding
coordinates on the part manifolds. If the user chooses object parts
from different images a cross-manifold optimization is carried out to
retrieve a single shape that blends together properties from different
images.

\begin{figure*}[t!]
    \centering
    \includegraphics[width=\textwidth]{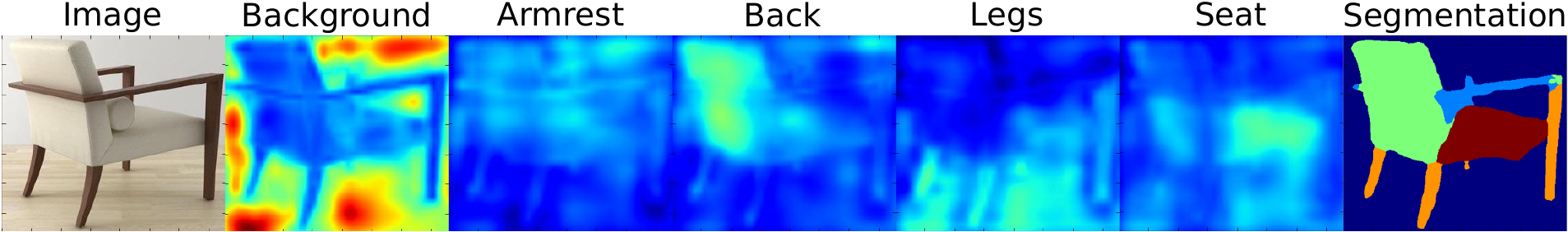}
    \caption{
	Example of the semantic segmentation performed by the first stages of our architecture. We can see the output probabilities for each of the parts and the background give a very strong prior of were the parts of an object can be found. Not requiring labels for each part in the input image makes our approach very easy to use and increases dramatically its applicability.
    }
    \label{fig:semSeg}
\end{figure*}

\noindent{\bf Training:} At training time, our method takes as input a
class-specific collection of 3D shapes (we used
$ShapeNet$~\cite{shapenet2015}) for which part label annotations are
available. The first step at training time is to \emph{learn a separate
  shape manifold for each object part} (see
Fig.~\ref{fig:part_manifold}). Each shape is represented with a Light
Field descriptor~\cite{LfDescriptor} and characterized with a pyramid
of HoG features. The manifolds are then built using non-linear
multi-dimensional-scaling (MDS) and the $L_2$ norm between feature
vectors as the distance metric -- in each resulting low-dimensional
manifold, objects that have similarly shaped parts are close to each
other. So far these manifolds of object parts (for instance
\emph{back-rests}, \emph{arm-rests}, \emph{legs}, \emph{seats} in the
case of chairs) contain 3D shapes. The second step at training time is
to train a CNN to embed images onto each part manifold by \emph{regressing
their coordinates}. We create a set of synthetic training images with
per pixel semantic segmentation annotations for the object parts and
ground truth manifold coordinates. The architecture of this novel CNN
(which we denote \textbf{ManifoldNet} and is shown in
Fig.~\ref{fig:manifold_network}) has three clear parts: a set of fully
convolutional layers for semantic segmentation of the object into
parts; a set of convolutional feature extraction layers; and a set of
fully connected layers for manifold coordinate regression. This
architecture can be trained end-to-end. We give an example of the
produced semantic segmentation in Fig.~\ref{fig:semSeg}.

\noindent{\bf Retrieval:} At test time, given a new query image of an
unseen object, \textbf{ManifoldNet} can embed it into each of the part
manifolds by regressing the coordinates. More importantly, our
retrieval system can take more than one image as input, picking
different object parts from each image. Note that \textbf{ManifoldNet}
only needs the input images and the name of the object part that will
be used from each image. The network learns jointly to segment the
image into parts and to regress the manifold coordinates and therefore
it does not require any manual annotations as input. A {\bf
  cross-manifold optimization} will then take the coordinates on each
of the part manifolds as input and return the coordinates of a unique
3D shape that blends the different object parts together. This is
achieved through an energy optimization approach, described in
section~\ref{sec:blending}.

\section{Methodology}

\subsection{Building shape manifolds for object parts}

We choose to create an embedding space that captures the similarity between the shape of object parts
based exclusively on the 3D shapes. The reason behind this choice is
that 3D models capture a more complete, pure and reliable
representation of geometry as opposed to images that often display
occlusions, or other distracting factors such as texture or shading
effects. We then rely on our new CNN architecture to map images onto
the same embedding by regressing their coordinates on the
corresponding manifolds.

\subsubsection{Defining a smooth similarity measure between 3D shapes}
\label{sec:optimised_hog_lfd}

Shape similarity between object parts is defined as follows. Given
a shape $S_i$, we define its Light-field Descriptor (LfD)~\cite{LfDescriptor} $L_i$ as the
concatenation of the HoG responses~\cite{Dalal_HoG} $L_i =
[H_1;H_2;...;H_k]$. The value of $k$ is fixed to $k=20$ throughout
this work. The light field descriptor $H_k$ for each view $k$ is
defined as $H_k = [H_k^{mid} ; H_k^{low}] \in \mathbb{R}^{2610}$. The
$L_2$ distance between feature vectors is then used as the similarity
measure between a pair of shapes $S_i$ and $S_j$: $d_{ij} =
\left|\left| L_i - L_j \right|\right|_2$ where $L_i \in \mathbb{R}^{52200}$. 
We found that using only the mid and low frequency parts of the HoG
pyramid leads to smoother transitions in shape similarity. For this reason we do not use the original 3 level HoG pyramids
but just the 2 higher levels of the pyramids. This allows for smooth transitions in shape similarity between parts making 
the shape blending possible. Due to most 3D models available in Internet datasets not being watertight but only polygon soups we are required to use
projective shape features like~\cite{LfDescriptor}. We now build separate manifolds for each object part. Each shape $S_i$ is
therefore split into its constituent parts $\forall \, S_i : \exists \, \{S_i^1;S_i^2;...;S_i^P\},$ 
where $P$ is the total number of parts and $S_i^p$ is the shape of
part $p$ of object $i$. If a part is not present in an object (for
instance, chairs without arm-rests) we set all the components of
the vector $L_i^p$ to zero, which is equivalent to computing the
HoG descriptor of an empty image.

\subsection{Building shape manifolds of parts}
\label{sec:building_manifolds}

\begin{figure*}[t]
    \centering
    \includegraphics[width=\textwidth]{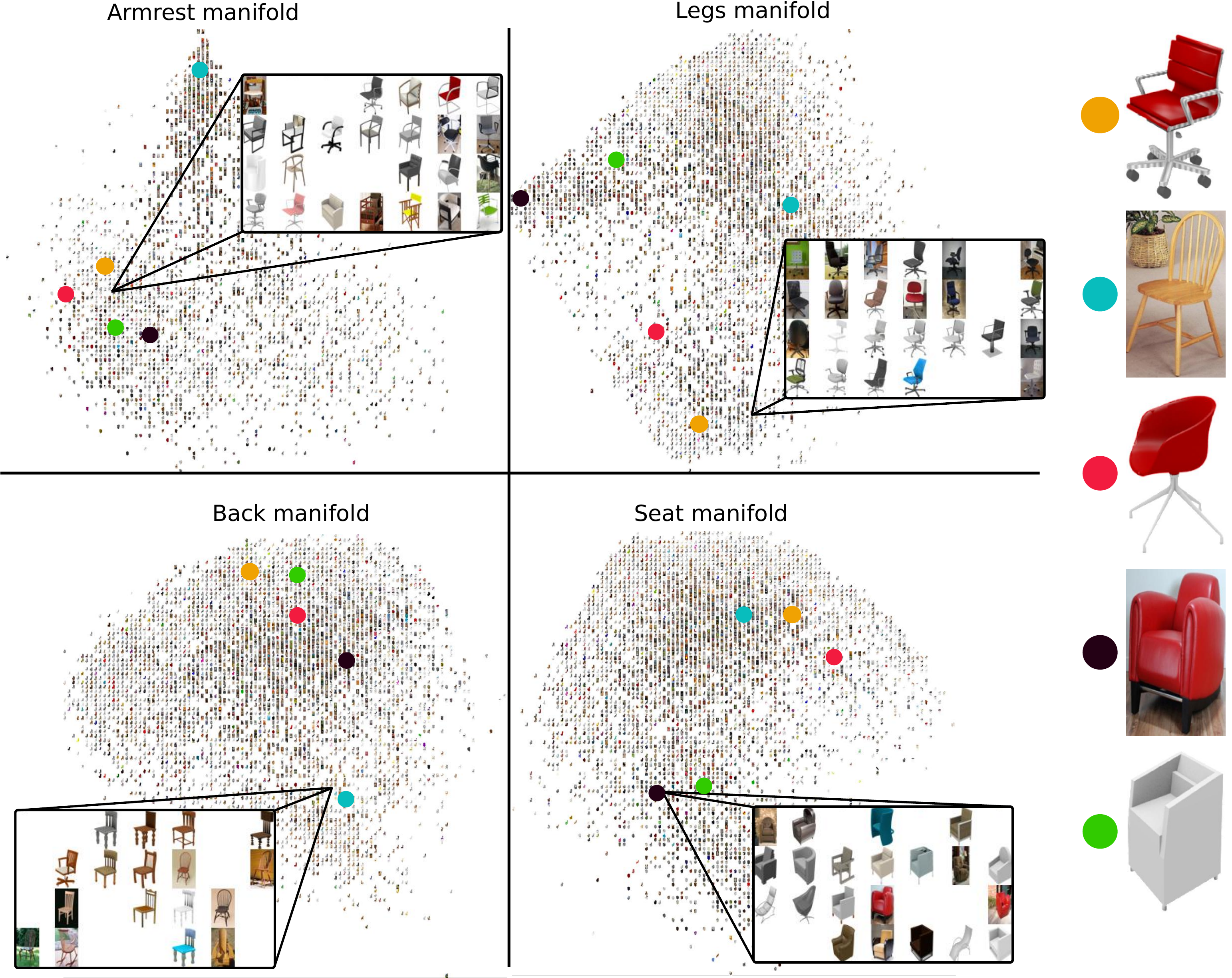}
    \caption{ Two dimensional visualizations of the four low dimensional
      part manifolds \emph{``backrest'', ``seat'', ``armrests''} and \emph{``legs''} for the chairs class. Probabilistic PCA
      has been used to provide a 2D visualization of the 128
      dimensional manifolds. Both images and 3D models have been
      represented in the manifolds. Objects with similarly shaped
      parts lie close to each other on the manifold. Several shapes
      are tagged in all four manifolds to show how vicinity changes
      for each part. All shapes and images exist in all manifolds.
      }
    \label{fig:manifold_coexistence}
\end{figure*}

Using the similarity measure between the shape of object parts 
we use it to construct a low dimensional representation
of the shape space. In principle, the original $L_i \in
\mathbb{R}^{52200}$ feature vectors could have been used to represent
each shape, since distances in that space reflect well the similarity
between shapes. We reduce the dimensionality from $52,200$
to $128$ dimensions and we use non-linear Multi-Dimensional Scaling
(MDS)~\cite{MDS_Kruskal1964} to build the shape manifolds.
We compute the distance matrix $D^p \in \mathbb{R}^{n \times n}$
as $D^p(i,j) = d^p_{ij}$, were $p$ is the index of the part and $n$ is
the total number of shapes. The manifold is built using MDS by
minimizing a Sammon Mapping error~\cite{Sammon_mapping_1969} defined
as

\begin{equation}
    E^p = \dfrac{1}{\sum_{i<j} D^p(i,j)} \sum_{i<j} \dfrac{(D^p(i,j)-D'^p(i,j))^2}{D^p(i,j)} ,
\end{equation}

where $D^p$ is the distance matrix between shapes in the original high
dimensional feature space $L^p$; and $D'^p$ is the distance matrix
between shapes in the new low dimensional manifold $L'^p$.
With the different manifolds $L'^p$ for each part $p$ computed, a low
dimensional representation of shape similarity exists and all 3D
shapes are already included in it. Adding new 3D shapes to the
manifold is done by solving an optimization that minimizes the
difference between the distances between all previous shapes and the
new shape in $D^p$ and the distances in $D'^p$ with respect to the
predicted embedding point. To understand the shape of the produced manifolds
we provide a 2D visualization in Fig.~\ref{fig:manifold_coexistence}. In this figure we
can better understand how the manifolds relate the different parts of an object.

\subsection{Learning to embed images into the shape manifolds}
\label{sec:learning_the_embedding}

Building the shape manifolds $L'^p$ for each part $p$ based only on
3D models, we have successfully abstracted away effects such as
textures, colours or materials. The next step is to train a deep
neural network that can map RGB images onto each manifold by
regressing the coordinates on each part manifold directly from RGB
inputs. Crucially, the input to the network must be simply the RGB
image and the name (label) of the object part $p$ selected from that
image -- for instance \emph{embed this image of a chair into the manifold of
{``chair legs''}}.

We propose a novel deep learning architecture, which we call
\textbf{ManifoldNet}, it performs three tasks: first,
it learns how to estimate the location of different parts in the image
by performing semantic segmentation, it then uses the semantic
labeling and the original input image to learn $p$ different
intermediate feature spaces for each object part and finally, $p$
different branches of fully connected layers will learn the final
image embedding into the respective part shape manifold. The network 
has a general core that performs semantic segmentation and specialized 
branches for each of the manifolds in a similar fashion as~\cite{ubernet}.

\begin{figure*}[t]
    \centering
    \includegraphics[width=\textwidth]{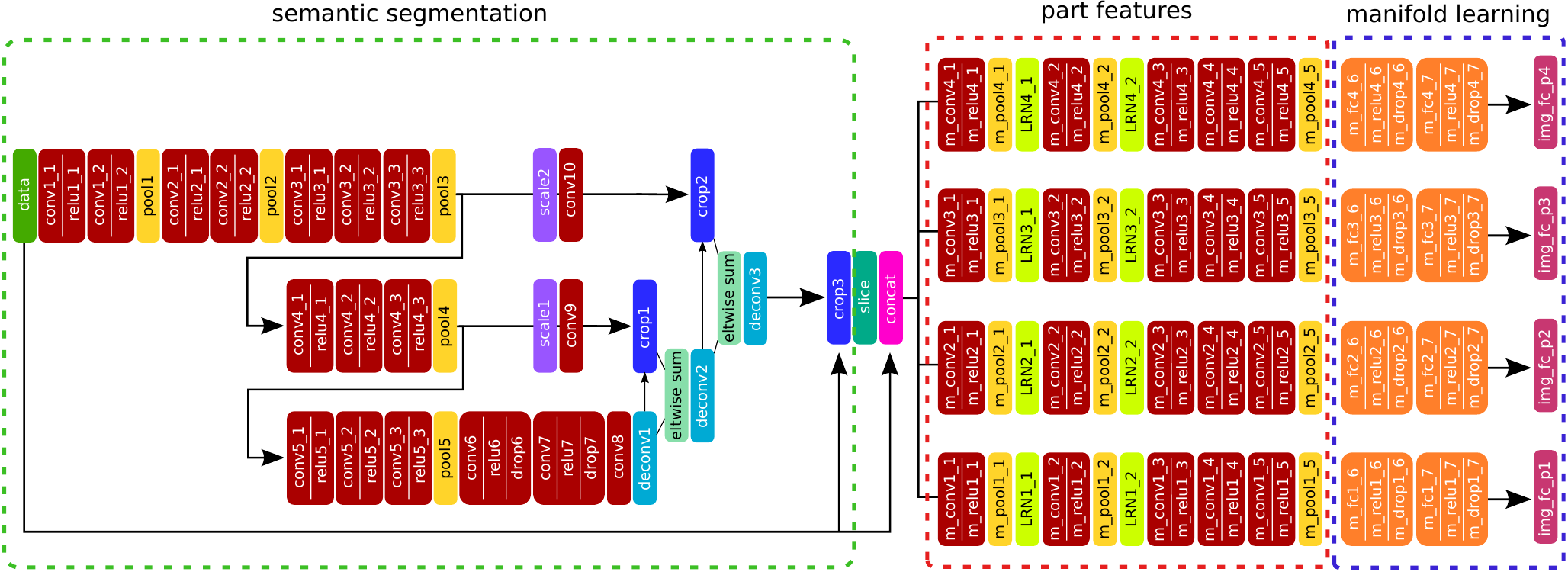}
    \caption{ Detailed \textbf{ManifoldNet} architecture. The first set of layers take
      care of assigning semantic part labels to each of the pixels in
      the image. The second stage extracts intermediate features to help the 
      manifold learning. The final stage uses a set of
      fully convolutional layers to regress the manifold coordinates.
    }
    \label{fig:detailed_architecture}
\end{figure*}

\subsubsection{\textbf{ManifoldNet}: A multi-manifold learning architecture}
\label{sec:manifold_net}

A summary of our new architecture is shown in
Fig.~\ref{fig:manifold_network} and a detailed description of all the
layers in Fig.~\ref{fig:detailed_architecture}. The common part of the
architecture, which performs the semantic segmentation, is a fully
convolutional approach closely related to~\cite{Long_2015_CVPR}. The
fully convolutional architecture uses a VGG-16
architecture~\cite{Simonyan15} for its feature layers. A combination
of FCN-32, FCN-16 and FCN-8 is used to obtain more detailed
segmentations but all sub-parts are trained together for simplicity.

The other two parts of the architecture shown in
Fig.~\ref{fig:manifold_network} take care of: \emph{(1)} creating an
intermediate feature space, and \emph{(2)} learning the manifold
embedding. The intermediate feature layers take as input the
concatenation of the original RGB image and the heat maps given as
output by the semantic segmentation layers to learn a feature
representation that eases the manifold learning task. Finally, the
manifold coordinate regression module is formed by 3 fully connected
layers (the first two use relu non-linearities). A dropout scheme is
used to avoid over-fitting to the data. Trained models and code 
are already public but due to anonymity we cannot disclose the URL.

\subsubsection{Details of the training strategy}
\label{sec:training_manifold_net}

The training of such a deep architecture requires careful attention. First, to avoid vanishing
gradients and to improve convergence, the semantic segmentation layers
are trained independently by using a standard cross-entropy classification loss:

\begin{equation}
L_{seg} = \dfrac{-1}{N} \sum_{n=1}^{N} \log{(p_{n,l_n})}
\end{equation}

where $p_{n,l_n}$ is the softmax output class probability for each pixel. 
A batch size of only 20 is used at this stage due to memory
limitations on the GPU and the high number of weights to be trained.

When trying to train the manifold layers we found out that convergence heavily
depended on big batch sizes and many iterations. At this point we used a learning
scheme that allowed us to have bigger batch sizes during training and faster computation
of each iteration. The trick is quite simple really, we precompute for all training 
images the output of the semantic layers and only train the part branches of the network.
By doing this we are training a substantially shallower network allowing for significantly 
bigger batch sizes. The network is trained by minimizing the following euclidean loss:

\begin{equation}
L_{mani} = \dfrac{1}{2N} \sum_{i=1}^{N} {\lVert x_{est}^i - x_{gt}^i \rVert}_2^2
\end{equation}

where $x_{est}^i$ are the manifold coordinates estimated by the
network and $x_{gt}^i$ are the ground truth manifold coordinates. The
Euclidean loss is chosen since the part shape manifolds are themselves
Euclidean spaces. With this good initialization of the weights we finally perform an end-to-end
training of all layers using only the final euclidean loss.

\noindent{\bf Training data and data augmentation:} the training images
are generated synthetically by rendering models
from $ShapeNet$~\cite{shapenet2015}. We use the 3D part annotations on
the 3D models, available from~\cite{Yi16}, to provide ground truth
values for the semantic segmentation. We generate 125 training images
per model from different poses, and a random RGB image taken from the
Pascal 3D~\cite{pascal3D_ref} dataset is added as background. 
To recap, the proposed approach is
invariant to pose and manages to learn solely from rendered synthetic images.

\subsection{Shape blending through cross-manifold optimization} 
\label{sec:blending}

Once the  manifold coordinates for the different object parts have
been estimated all the information needed to blend them into a single
3D model is available. We formulate this as a 3D shape retrieval problem: 
\emph{``find the 3D model, from the existing shapes represented in the
manifolds, that best fits the arrangement of parts''}. The user selects 
two (or more) images (or 3D models) and indicates the
part they wish to select from each one (note that no annotations are
needed, only the name/label of the part). The cross-manifold
optimization now finds the 3D shape in the collection that minimizes
the sum of the distances to each of the parts. In more detail -- first, all manifolds need to be normalized to allow
a meaningful comparison of distances. Then, given the set of manifold
coordinates for the selected parts, a shape prediction $b$ can be
defined as the concatenation of the respective part coordinates $b =
\{b^1;...;b^m\}$. The goal is now to retrieve a 3D model from the
shape collection whose coordinates $a = \{a^1;...;a^m\}$ are closest
to this part arrangement by minimizing the following distance:

\begin{equation}
B = \min_{a \in \mathbb{S}} \sum_{k=1}^{m} \lVert {a^{m} - b^{m}} \rVert
\end{equation}

where $\mathbb{S}$ is the set of existing shapes. Note that not all
parts need to be selected to obtain a blended shape, we define $m$ as
the subset of parts to be blended, where $m \subseteq p$. Also, notice
that blending can be done by combining any number of parts from any
number of sources (shapes/images).

\section{Results}

We perform a set of qualitative and quantitative experiments to
evaluate the performance of our approach. Allthough our approach performs
shape blending from several inputs this can be understood as a retrieval task.

\subsection{Quantitative results on Image-based Shape Retrieval}

The proposed approach will be at a disadvantage when trying to retrieve whole shapes and the same will 
happen when approaches that model the object as a whole try to retrieve parts. What has been have done 
is a experimental comparison of both approaches on both tasks, as a clean unbiased comparison cannot 
be done on a single experiment both approaches will be used to solve both tasks. This is possible as
all compared approaches can be used as a similarity measure between images and 3D models. By doing 
this, how much is lost can be measured when either modeling the whole object or the individual parts.

\subsubsection{Image-based Whole Shape Retrieval}

\begin{figure}[t!]
    \centering
    \includegraphics[width=\textwidth, trim={3mm 5mm 3mm 0mm}]{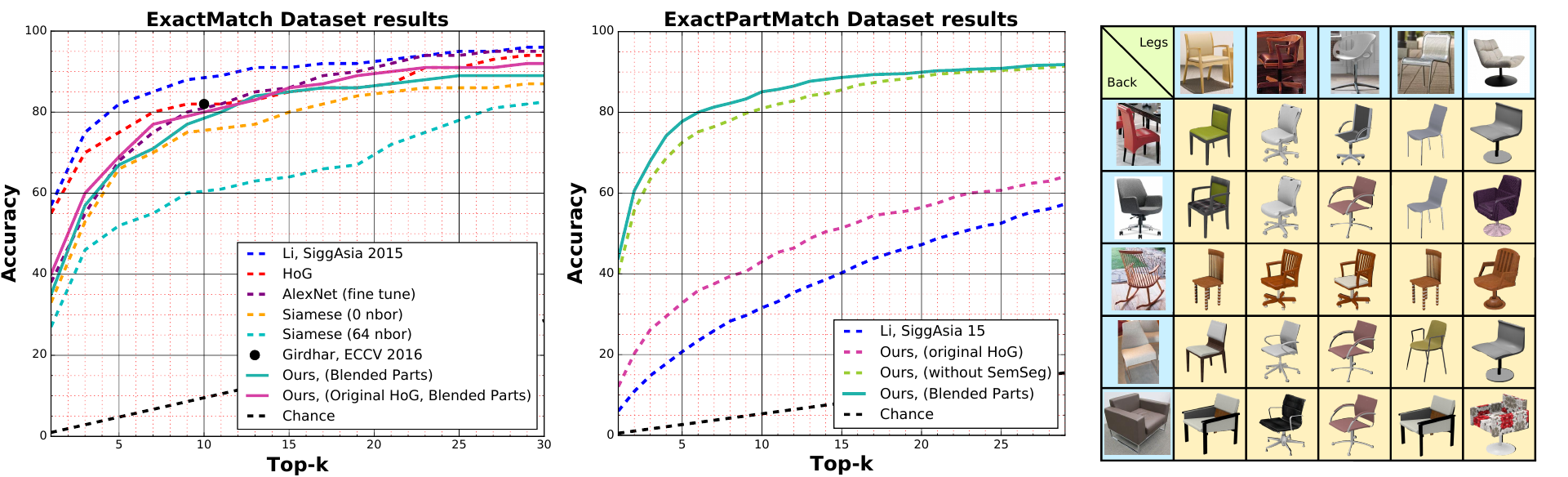}
    \caption{
      \textbf{Left}: Image-based whole shape retrieval results obtained on the ExactMatch dataset. 
      \textbf{Center}: Shape blending and retrieval from image-based part descriptions on the ExactPartMatch dataset. 
      \textbf{Right}: Matrix showing combination results. Experiment performed by generating all possible combinations of legs and backs from 10 test shapes.
      }
    \label{fig:mainResults}
\end{figure}

We perform whole shape image-based retrieval on the \textbf{ExactMatch} dataset~\cite{li2015jointembedding}. The 
experiment compares against Li~\cite{li2015jointembedding}, a state-of-the-art deep volumetric approach in 
Girdhar~\cite{Girdhar16b}, HoG, AlexNet and Siamese networks. Two versions of the proposed approach using 
the original three level HoG pyramid features to build the manifold and the two level HoG manifold features 
that have been shown to be better fitted for a smoother shape similarity measure. Our approach predicts 
the part coordinates separately in each of the part manifolds. The estimations of each part are then used 
to solve the blending optimization and obtain a single shape prediction. The fact that the neural network 
estimates the coordinates individually means that all the part co-occurrence information that is implicitly 
encoded in the approaches that model the object as a whole is lost during training, nevertheless, the 
proposed approach can still yield good results that are comparable to those that model the whole 
shape. It can also be seen that for the exact shape retrieval the original three level pyramid HoG 
features perform better which is to be expected. The results of these experiments are included in 
Fig.~\ref{fig:mainResults} (first image).

\subsubsection{Image-based Part Shape Blending}

To test the performance of part blending and retrieval a new dataset has 
been created using a subset of the shapes from the Shapenet database~\cite{shapenet2015} and the images from the 
ExactMatch dataset~\cite{li2015jointembedding} to create the \textbf{Exact\textit{Part}Match} 
dataset. The task is to find the correct 3D shape out of all the annotated 3D models using the parts from the specified inputs. As in Shapenet
many of the 3D models are repeated many times (e.g. ikea chairs) we need to control that there is only one correct match with the part mix.
The dataset contains \textbf{845 test cases}. Each test case if assigned to one of 187 hand annotated ground truth 3D models, the same part shape 
combination is tested using images from different angles, textures and lighting conditions. Each \textbf{test case} 
is a combination of parts from two different images for which a 3$D$ model 
exists in Shapenet. There are examples of all possible part combinations. The whole dataset is already 
public online, the link is not disclosed due to anonymity. The results of these experiments are included in Fig.~\ref{fig:mainResults} (second 
image). Also in Fig.~\ref{fig:mainResults} (third image) we include an experiment that shows the 
product of crossing 5 leg samples with 5 back samples densely, this shows the kind of results to 
expect from our approach.

\begin{figure*}[t!]
    \centering
    \includegraphics[width=\textwidth,trim={0mm, 0mm, 0mm, 0mm}]{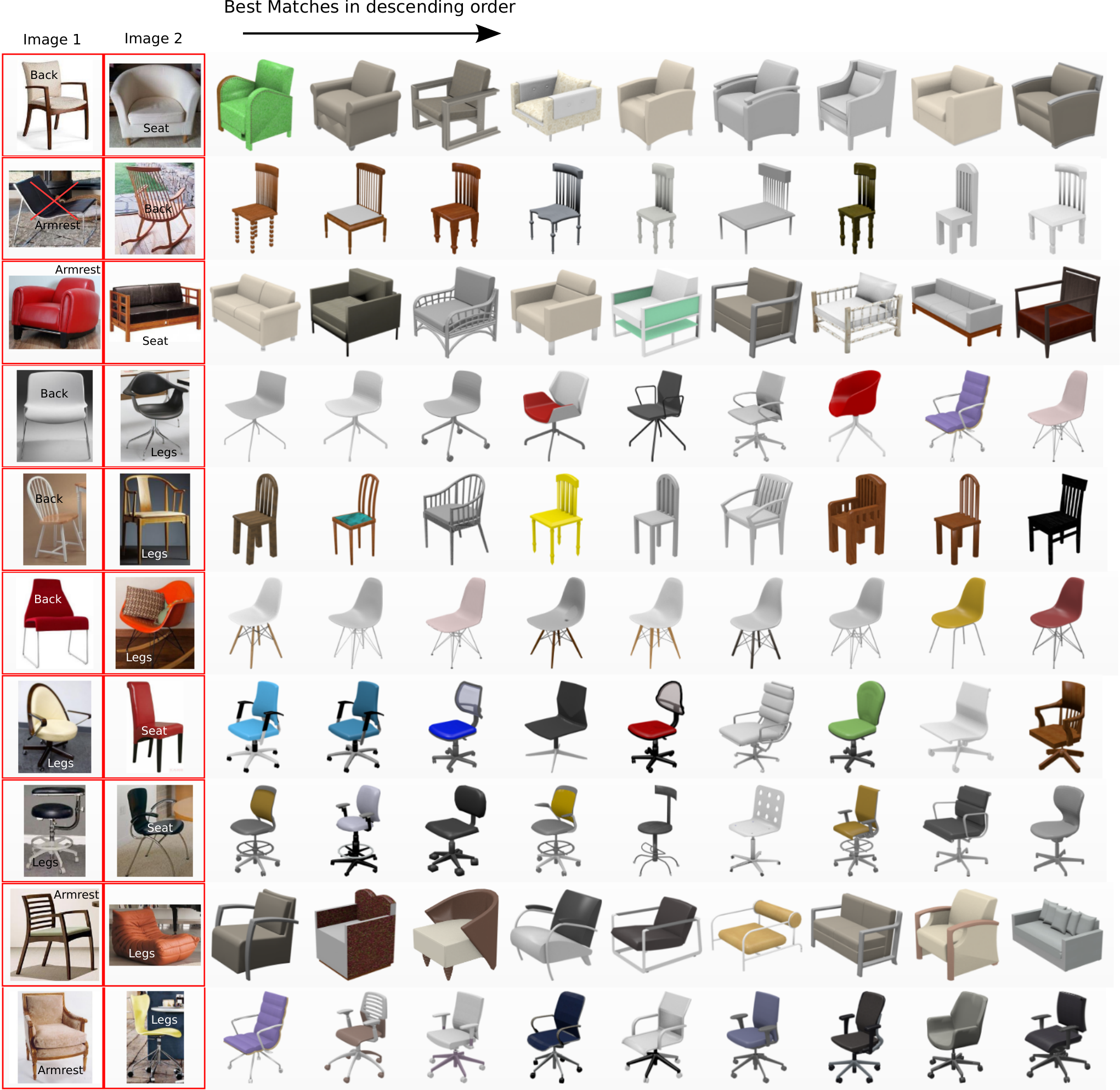}
    \caption{
	Qualitative results extracted to show performance on different part arrangements using parts from two different image inputs. The results show the closest matching shapes in the Shapenet dataset. If the part is not present, like in the second row, an $X$ in the colour of the desired part is used to label the non-presence of that part.
    }
    \label{fig:qualitativeResults}
\end{figure*}

\begin{figure*}[t!]
    \centering
    \includegraphics[width=\textwidth,trim={0mm, 0mm, 0mm, 0mm}]{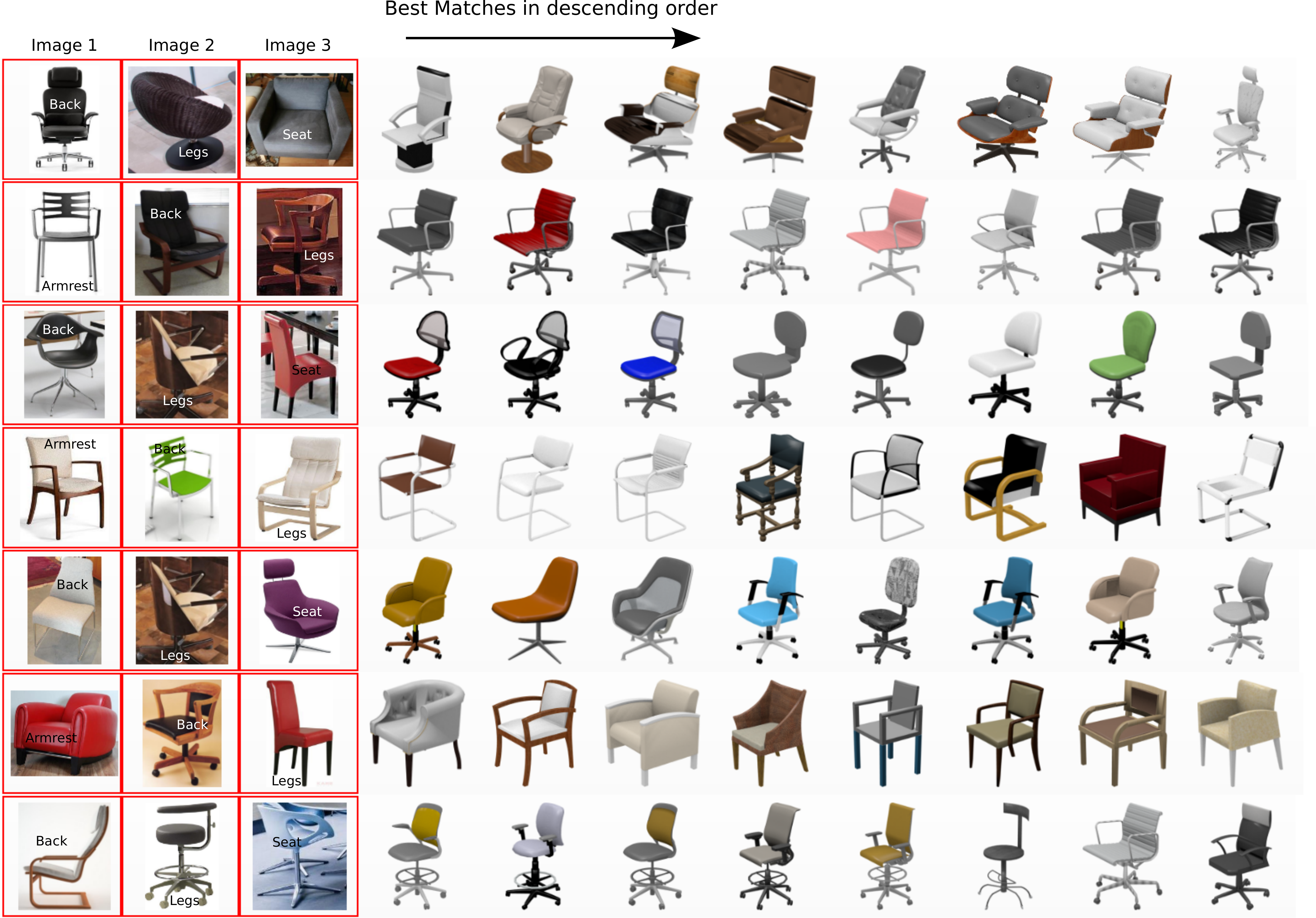}
    \caption{
	Qualitative results extracted to show performance on different part arrangements using parts from three different image inputs. The results show the closest matching shapes in the Shapenet dataset.
    }
    \label{fig:qualitativeResults3shapes}
\end{figure*}

For all approaches similarity is estimated in shape manifold space and then the multi-manifold 
optimization is performed, which is required to obtain the most similar shape. The approaches 
shown are $Ours$, with and without semantic segmentation, $Ours$ with semantic segmentation but 
using the original three level HoG features, $Li's SiggAsia 15$~\cite{li2015jointembedding} and 
random chance. ~\cite{li2015jointembedding} struggles to get results as good as the ones obtained 
by modeling the parts. Their holistic representation enables them to better model a whole object 
but loose substantial performance when trying to identify individually the parts. In contrast our 
approach looses the information of part co-occurrence in favour of being capable to model the 
parts individually. Also, exploiting the semantic segmentation of the input is 
consistently better as it defines the actual interest zones in the image. If considering top-5 results without semantic segmentation $65\%$ is obtained but when using semantic segmentation $76\%$ is 
obtained, which is a substantial $11\%$ improvement in performance. It can also be seen that trying to blend shapes when using shape similarities 
that do not correctly model smoothness over shape has a tremendous impact in performance (Ours\_original\_HoG).

\subsection{Qualitative results on Image-based Shape Retrieval}

To further asses the quality of the results examples are shown depicting the performance of the approach being applied 
to real images. In Fig.~\ref{fig:qualitativeResults} and in Fig.~\ref{fig:qualitativeResults3shapes} many example images 
taken from the \textbf{Exact\textit{Part}Match} dataset detailed in previous sections are shown. Our approach searches over the 
entirety of Shapenet to find the closest matches to the input part arrangement. Many different part 
arrangements are accounted for in the figure to show that the proposed approach can capture not only the big differences 
but also more subtle differences like the number of legs in the base of a specific swivel chair, the fact that a wooden 
chair with a back made of bars has a round top or a flat top, capturing the detailed shape of the interconnecting supports of the four legs, etc.

\section{Conclusions}

An approach capable of modelling and blending object parts from images and 3D models has been presented. This approach has 
demonstrated that by using a common manifold representation very elaborate queries can be done in massive Internet 
databases. It has also shown to be capable to produce accurate shape retrieval which proves its understanding of the underlying shapes.
This provides a natural link between shape and image datasets and opens numerous possibilities of 
application to similar tasks. Also, by understanding the parts that give semantic meaning to an object volumetric
approaches like~\cite{choy20163d,conf/cvpr/KarTCM15,qi2016volumetric,Haoqiang_CVPR17} could potentially address their difficulties to
produce details in their volumetric estimations by either enforcing the shape of the part explicitly or balancing the voxel occupancy probabilities
for each part independently.


\end{document}